\pdfoutput=1
\documentclass[11pt]{article}
\usepackage[affil-sl,noblocks]{authblk}

\usepackage{booktabs}

\usepackage{csquotes}
\usepackage{mdframed}

\usepackage[final]{acl}

\usepackage{times}
\usepackage{latexsym}

\usepackage[T1]{fontenc}

\usepackage[utf8]{inputenc}

\usepackage{microtype}

\usepackage{inconsolata}

\usepackage{graphicx}

\usepackage[absolute,showboxes,overlay]{textpos}
\TPMargin{3pt}

\title{Detecting Sexism in German Online Newspaper Comments\\with Open-Source Text Embeddings\\(Team GDA, GermEval2024 Shared Task 1: GerMS-Detect,\\Subtasks 1 and 2, Closed Track)}

\author[1]{Florian Bremm}
\author[1]{Patrick Gustav Blaneck}
\author[2]{Tobias Bornheim}
\author[1,3]{Niklas Grieger}
\author[1,*]{Stephan Bialonski}
\affil[1]{Department of Medical Engineering and Technomathematics\authorcr Institute for Data-Driven Technologies, FH Aachen University of Applied Sciences, Jülich, Germany}
\affil[*]{\textit{bialonski@fh-aachen.de}\authorcr}
\affil[2]{ORDIX AG\authorcr}
\affil[3]{Department of Information and Computing Sciences, Utrecht University, Utrecht, The Netherlands\authorcr}

\date{}

\begin{document}
\maketitle
\begin{abstract}
    Sexism in online media comments is a pervasive challenge that often manifests subtly, complicating moderation efforts as interpretations of what constitutes sexism can vary among individuals.
    We study monolingual and multilingual open-source text embeddings to reliably detect sexism and misogyny in German-language online comments from an Austrian newspaper.
    We observed classifiers trained on text embeddings to mimic closely the individual judgements of human annotators.
    Our method showed robust performance in the \emph{GermEval 2024 GerMS-Detect} Subtask 1 challenge, achieving an average macro F1 score of 0.597 (4th place, as reported on Codabench).
    It also accurately predicted the distribution of human annotations in GerMS-Detect Subtask 2, with an average Jensen-Shannon distance of 0.301 (2nd place).
    The computational efficiency of our approach suggests potential for scalable applications across various languages and linguistic contexts.
\end{abstract}

\begin{textblock*}{20cm}[0.5,0](10.5cm,27.8cm)
  \centering
  \small
  This work was peer-reviewed and published in the Proceedings of GermEval 2024, available at \url{https://aclanthology.org/2024.germeval-2.5/}. 
  Please cite as: Bremm F., Blaneck P.G., Bornheim T., Grieger N., \& Bialonski, S. "Detecting Sexism in German Online Newspaper Comments with Open-Source Text Embeddings (Team GDA, GermEval2024 Shared Task 1: GerMS-Detect, Subtasks 1 and 2, Closed Track)." Proceedings of GermEval 2024 Task 1 GerMS-Detect Workshop on Sexism Detection in German Online News Fora (GerMS-Detect 2024). Association for Computational Lingustics, 2024.
\end{textblock*}

\section{Introduction}

The reliable detection of sexism and misogyny in online discussions has received increased attention in recent years~\citep{Fontanella2024}.
Since the events of \enquote{Gamergate} in August 2014~\citep{Massanari2016}, a harassment campaign targeting female journalists, research on sexism and misogyny in online platforms has gained momentum.
Exposure to sexism can have tangible negative effects, for example discouraging women from participating in online discussions, as shown by an online survey conducted by an Austrian newspaper~\citep{Krenn2024}.
Given the often subtle and subjective nature of sexist content, moderators face significant challenges in identifying it.
This highlights the need for effective detection tools that can support moderators creating more inclusive online spaces for women.

Previous efforts to automate the detection of sexism and misogyny have typically relied on machine learning methods, often treating sexism and misogyny as forms of hate speech~\citep{Jahan2023}.
Numerous datasets have been created to support the training and validation of general hate speech detection models~\citep{Poletto2021,Yu2024}.
More recently, specialized datasets aimed specifically at sexism or misogyny detection have been released for different languages, including English~\citep{Anzovino2018}, Spanish~\citep{Rodriguez2020}, and French~\citep{Chiril2020}.
Alongside these developments, several competitions, such as SemEval-2019 Task 5~\citep{Basile2019}, EXIST 2022~\citep{Rodriguez2022}, and SemEval-2023 Task 10~\citep{Kirk2023}, have been held to promote progress in identifying sexism and misogyny.
However, German-language resources for sexism detection have been particularly limited~\citep{Yu2024}.
The introduction of GERMS-AT~\citep{Krenn2024}, a dataset of about 8000 online comments of an Austrian newspaper annotated for sexist content, has significantly improved the prospects for developing and evaluating sexism detection models in German.
This dataset includes diverse annotations from multiple individuals, capturing the variability in human judgment.

In this contribution, we study the ability of open-source text embedding models, i.e., the multilingual \enquote{mE5-large}\footnote{\url{https://huggingface.co/intfloat/multilingual-e5-large}}~\citep{Wang2024a} and the monolingual \enquote{German BERT large paraphrase cosine}\footnote{\url{https://huggingface.co/deutsche-telekom/gbert-large-paraphrase-cosine}} model, to reliably detect sexism and misogyny in German-language online comments (GERMS-AT).
Using these text embeddings, we observed that traditional machine learning classifiers, which are fast and inexpensive to train, robustly predict the judgments of human annotators.
We detail our approach and describe the models that were evaluated in the \emph{GermEval 2024 GerMS-Detect} shared tasks and the results we obtained on out-of-sample data.
The implementation details of our experiments are available online\footnote{\url{https://github.com/dslaborg/germeval2024}}.

\section{Data and Tasks}
\label{sec:data_task}

\subsection{Data}
\label{ssec:data}

The dataset of the \emph{GermEval 2024 GerMS-Detect} Shared Task consisted of 7984 German-language comments from the comment section of an Austrian online newspaper~\cite{Krenn2024}. 
While all comments were in German, many comments contained Austrian dialects or slang (see Figure~\ref{fig:example_annotation}, Example 1).

\begin{figure}[htbp]
    \centering
    \begin{mdframed}[
            leftmargin=0pt,
            rightmargin=0pt,
        ]
        \small
        \textbf{Example 1:}

        \enquote{Des Oaschloch is eh scho berühmt, de virz'g Jungfrauen oide, notgeile Nonnen.}
        {\tiny(ID: \texttt{e1e80ff680f874d49ddfe33ac846a454})}

        {\tiny{Trans.: \enquote{This asshole is already famous anyway, the forty virgins, old, horny nuns.}}}

        \vspace{0.5em}

        \begin{tabular}{lll}
            No. \emph{0-absence}: & 1 & {\tiny(A001)}                         \\
            No. \emph{1-mild}:    & 0 &                                       \\
            No. \emph{2-present}: & 1 & {\tiny(A007)}                         \\
            No. \emph{3-strong}:  & 5 & {\tiny(A002, A003, A004, A005, A012)} \\
            No. \emph{4-extreme}: & 3 & {\tiny(A008, A009, A010)}             \\
        \end{tabular}

        \vspace{0.5em}

        \textbf{Example 2:}

        \enquote{Warum wählen dann aber immer noch 36\% der Frauen in Österreich die övp?}
        {\tiny(ID: \texttt{0917bc805a3b4c3086ee7101f2740dad})}

        {\tiny{Trans.: \enquote{Then why do 36\% of women in Austria still vote for the ÖVP?}}}

        \vspace{0.5em}

        \begin{tabular}{lll}
            No. \emph{0-absence}: & 4 & {\tiny(A002, A009, A010, A012)} \\
            No. \emph{1-mild}:    & 0 &                                 \\
            No. \emph{2-present}: & 0 &                                 \\
            No. \emph{3-strong}:  & 0 &                                 \\
            No. \emph{4-extreme}: & 0 &                                 \\
        \end{tabular}

    \end{mdframed}
    \caption{
        Comments from the provided training dataset with annotations grouped by label (annotators shown in parentheses).
        The comment in \emph{Example 1} contains an Austrian dialect and was annotated by all ten experts receiving a variety of labels.
        \emph{Example 2} was only annotated by four experts and received the same label from all of them.
    }
    \label{fig:example_annotation}
\end{figure}

\begin{figure}[htbp]
    \centering
    \includegraphics[width=\linewidth]{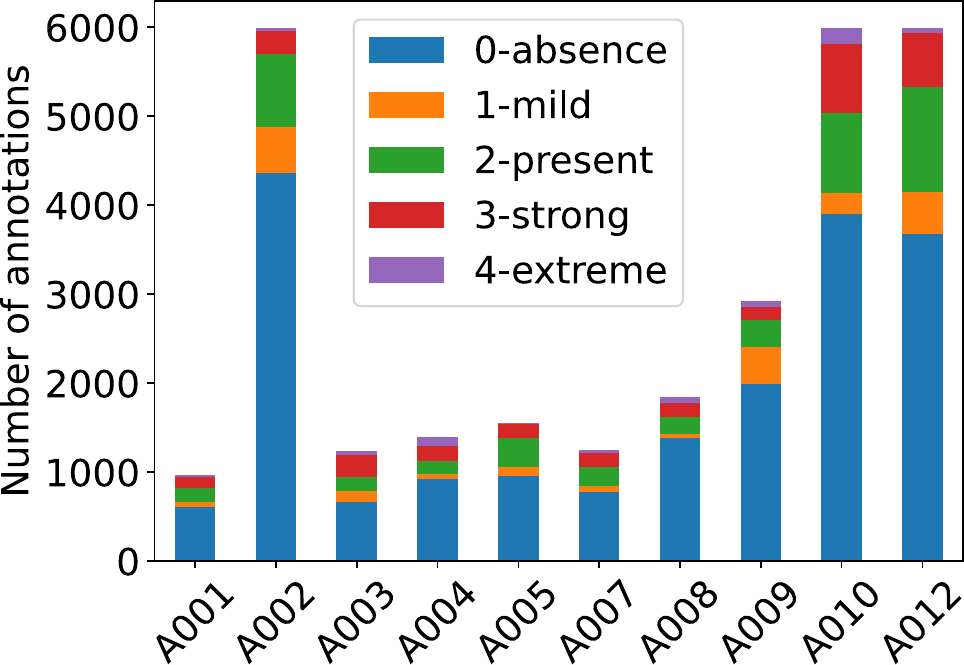}
    \caption{
        Distribution of the labels assigned by each annotator (A001--A012).
        Note that there are no annotations from users A006 and A011.
    }
    \label{fig:dist_annotations}
\end{figure}

Each comment was annotated by at least four annotators out of a group of ten human experts following specific guidelines\footnote{\url{https://ofai.github.io/GermEval2024-GerMS/guidelines.html}~\citep{Krenn2024}}.
Annotators were asked to label each comment as either not-sexist (0) or sexist (1--4).
If a comment was identified as sexist, annotators assigned a label between 1 and 4, indicating the severity of the sexism or misogyny (1-mild, 2-present, 3-strong, 4-extreme).
The annotations were highly subjective and varied significantly between annotators (see Figure~\ref{fig:example_annotation}).
Figure~\ref{fig:dist_annotations} illustrates this variability in the annotations and highlights the imbalance in the number of comments labeled by each annotator.
Additionally, Figure~\ref{fig:dist_annotations} shows, that the distribution of the five labels is highly imbalanced, with the majority of comments being labeled as not-sexist (0) by all annotators.

For the final phase of the Shared Task, the organizers provided a training dataset containing 5998 comments (75.1\%) and a test dataset containing 1986 comments (24.9\%), which was used for the evaluation of the final models on the competition website.
The test dataset did not contain any annotations, but included the IDs of the annotators who labeled each comment.
We employed two different data splits for model exploration and final training, respectively.
During model exploration, we randomly split the provided training dataset into a smaller training set with 80\% of the comments and a validation set containing the remaining 20\%.
We then created annotator-specific training sets by filtering the reduced training set for the annotations of each annotator.
Furthermore, each annotator-specific training set was split into five folds for a cross-validation setup.
For final training, we used the entire provided training set and, similar to the model exploration phase, created annotator-specific training sets with 10\% of the data reserved for early stopping.

\subsection{Tasks}
\label{ssec:tasks}

The Shared Task consisted of two subtasks.
In \emph{Subtask 1}, all annotations of a comment were aggregated into a single prediction target using various strategies.
The goal was then to predict the aggregated label of each aggregation strategy for each comment.
The following aggregation strategies were used:
\begin{itemize}
    \item \emph{Majority (binary prediction target)}:
          A comment is labeled as \emph{not-sexist} or \emph{sexist} depending on whether the majority of annotators labeled the comment as not-sexist (0) or sexist (1--4).
          If there is no majority, both labels are considered valid.
          In Figure~\ref{fig:example_annotation}, \emph{Example 1} would be labeled as \emph{sexist} and \emph{Example 2} as \emph{not-sexist}.
    \item \emph{One (binary)}:
          If at least one annotator labeled a comment as sexist (1--4), the comment is labeled as \emph{sexist}, otherwise as \emph{not-sexist}.
          In Figure~\ref{fig:example_annotation}, \emph{Example 1} would be labeled as \emph{sexist} and \emph{Example 2} as \emph{not-sexist}.
    \item \emph{All (binary)}:
          If all annotators labeled a comment as sexist (1--4), the comment is labeled as \emph{sexist}, otherwise as \emph{not-sexist}.
          In Figure~\ref{fig:example_annotation}, both examples would be labeled as \emph{not-sexist}.
    \item \emph{Majority (multi-class)}:
          The label of a comment is the majority label of all annotators for this comment.
          If there is no majority, each of the labels assigned by the annotators is considered valid.
          In Figure~\ref{fig:example_annotation}, \emph{Example 1} would be labeled as \emph{3-strong} and \emph{Example 2} as \emph{0-absence}.
    \item \emph{Disagreement (binary)}:
          If at least one annotator labeled a comment as sexist (1--4), while at least one other annotator labeled the same comment as not-sexist (0), the comment is labeled as \emph{disagreed}.
          Otherwise, the comment is labeled as \emph{agreed}.
          In Figure~\ref{fig:example_annotation}, \emph{Example 1} would be labeled as \emph{disagreed} and \emph{Example 2} as \emph{agreed}.
\end{itemize}

In contrast to the binary and multi-class targets of \emph{Subtask 1}, the goal of \emph{Subtask 2} was to model the relative distribution of annotations per comment.
The following two distributions were of interest:
\begin{itemize}
    \item \emph{Binary}:
          The portion of annotators labelling a comment as \emph{not-sexist} (0) or \emph{sexist} (1--4) respectively.
          In Figure~\ref{fig:example_annotation}, \emph{Example 1} would be labeled as 10\% \emph{not-sexist} and 90\% \emph{sexist}.
          \emph{Example 2} would be labeled as 100\% not-sexist.
    \item \emph{Multi-Class}:
          Each prediction target represents the portion of annotators labelling a comment as one of the five labels (0--4).
          In Figure~\ref{fig:example_annotation}, \emph{Example 1} would be labeled as 10\% \emph{0-absence}, 0\% \emph{1-mild}, 10\% \emph{2-present}, 50\% \emph{3-strong}, and 30\% \emph{4-extreme}.
          \emph{Example 2} would be labeled as 100\% \emph{0-absence}.
\end{itemize}

\section{Methods and Results}
\label{sec:methods_results}

Both subtasks of the \emph{GermEval 2024 GerMS-Detect} Shared Task required knowledge of the distribution of annotations per comment.
Since the test dataset contained information about the annotators that labeled the comments, we decided to train individual models for each annotator.
In our approach, we combined pre-trained open-source large language models for text embeddings with simple classifiers to predict the annotations of an annotator.

\subsection{Model Architecture}

All comments were embedded into high-dimensional vector spaces using either %
(i) the \enquote{German BERT large paraphrase cosine} (GBERT-large-pc) model\footnote{\url{https://huggingface.co/deutsche-telekom/gbert-large-paraphrase-cosine}}, a version of the monolingual German-BERT model \citep{Chan2020} that was fine-tuned for text embeddings by Deutsche Telekom or %
(ii) the multilingual \enquote{mE5-base}\footnote{\url{https://huggingface.co/intfloat/multilingual-e5-base}} or \enquote{mE5-large}\footnote{\url{https://huggingface.co/intfloat/multilingual-e5-large}}~\citep{Wang2024a} models.
A general overview of the models used for text embeddings is provided in Table~\ref{tab:models}.

\begin{table}[htbp]
    \centering
    \begin{tabular}{lcc}
        \toprule
        Model          & Layers & Embedding Size \\
        \midrule
        mE5-base       & 12     & 768            \\
        mE5-large      & 24     & 1024           \\
        GBERT-large-pc & 24     & 1024           \\
        \bottomrule
    \end{tabular}
    \caption{
        Overview of the models used for text embeddings.
        The models were used as is, without any further fine-tuning.
    }
    \label{tab:models}
\end{table}

The resulting text embeddings were then used as input features for each of the following classifiers: (i) Multilayer Perceptron (MLP), (ii) Random Forest (RFC), and (iii) Support Vector Machine (SVC).

\subsection{Model Training and Evaluation}

Model training consisted of a model exploration phase, where we optimized hyperparameters and selected the best-performing classifier, and the final training, where we retrained the optimal training configurations on the entire training set for submission to the competition.
During training, only the classifier was updated, while the text embeddings were kept fixed.
When training the Multilayer Perceptrons, we used 10\% of the training data as an early stopping set to prevent overfitting.

We optimized hyperparameters separately for each annotator and classifier type using 5-fold cross-validation on each annotator's training dataset (see Section~\ref{ssec:data} for details on the data split).
Table~\ref{tab:hyperparameters} provides an overview of the hyperparameters of each classifier and their search ranges that were explored using grid search.
To mitigate the class imbalance in the dataset (see Section~\ref{ssec:data}), we balanced the classes for each annotator-specific dataset by either using class weights (preferred) or oversampling, depending on the respective model implementation by scikit-learn\footnote{https://scikit-learn.org/stable/} (see Table~\ref{tab:hyperparameters}).

\begin{table}[tbp]
    \centering
    \begin{tabular}{lc}
        \toprule
        Hyperparameter       & Value/Range                  \\
        \midrule
        \multicolumn{2}{l}{\textbf{Multilayer Perceptron}}  \\
        hidden\_layer\_sizes & $[64, 2048]$                 \\
        class\_weight        & oversampled                  \\
        \midrule
        \multicolumn{2}{l}{\textbf{Random Forest}}          \\
        n\_estimators        & $[10, 560]$                  \\
        criterion            & gini                         \\
        max\_depth           & $[1, 91]$                    \\
        class\_weight        & balanced                     \\
        \midrule
        \multicolumn{2}{l}{\textbf{Support Vector Machine}} \\
        C                    & $[1, 91]$                    \\
        kernel               & rbf                          \\
        class\_weight        & balanced                     \\
        \bottomrule
    \end{tabular}
    \caption{
        Overview of the hyperparameter ranges used for tuning the classifiers.
        The hyperparameters were explored using grid search with 5-fold cross-validation.
        The hyperparameter \enquote{class\_weight} indicates how the class imbalance was addressed.
    }
    \label{tab:hyperparameters}
\end{table}

After hyperparameter tuning, we retrained models for each annotator in the best-performing configuration on the entire annotator-specific training set and evaluated them on the validation set to identify the best-performing classifier type.
For evaluation, we aggregated the predictions of each annotator's model using the respective aggregation strategies for Subtasks 1 and 2 (see Section~\ref{ssec:tasks}).
We then evaluated the performance of the aggregations using the Macro-F1 score for Subtask 1 and the Jensen-Shannon distance~\citep{Lin1991} for Subtask 2.

Once the best-performing classifier was identified, we recombined the training and validation sets and retrained the models on the entire dataset for each annotator.
The final models then predicted the annotations of the test set, which were aggregated for Subtasks 1 and 2 using the same strategies as during model exploration and submitted to the competition.

\begin{figure}[htbp]
    \centering
    \includegraphics[width=\linewidth]{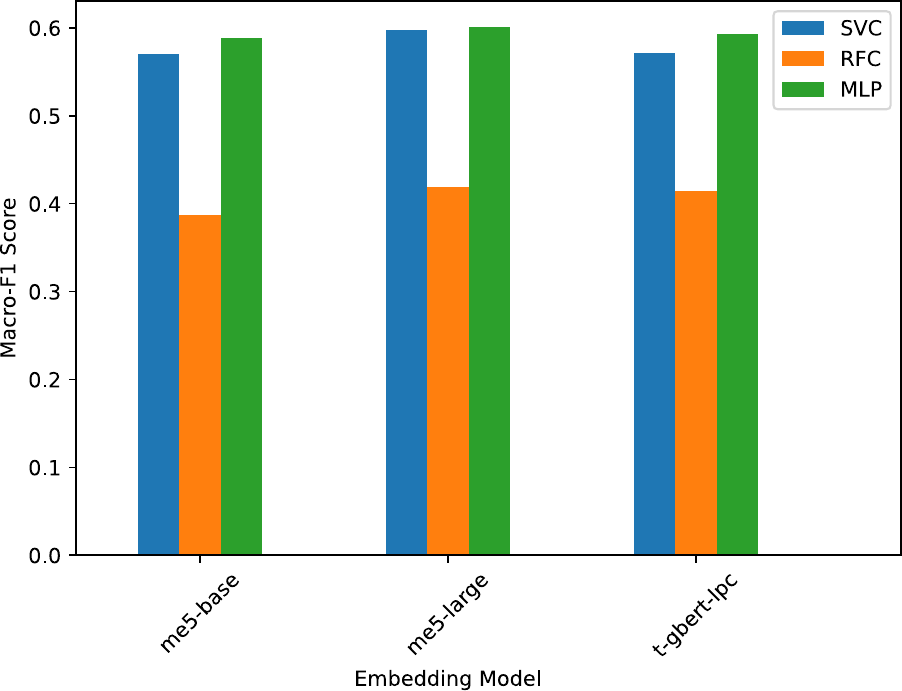}
    \caption{
        Macro-F1 scores our models achieved when aggregating the predictions for Subtask 1 on the validation set (higher is better).
    }
    \label{fig:validation_subtask1}
\end{figure}
\subsection{Results}

\begin{figure}[tbp]
    \centering
    \includegraphics[width=\linewidth]{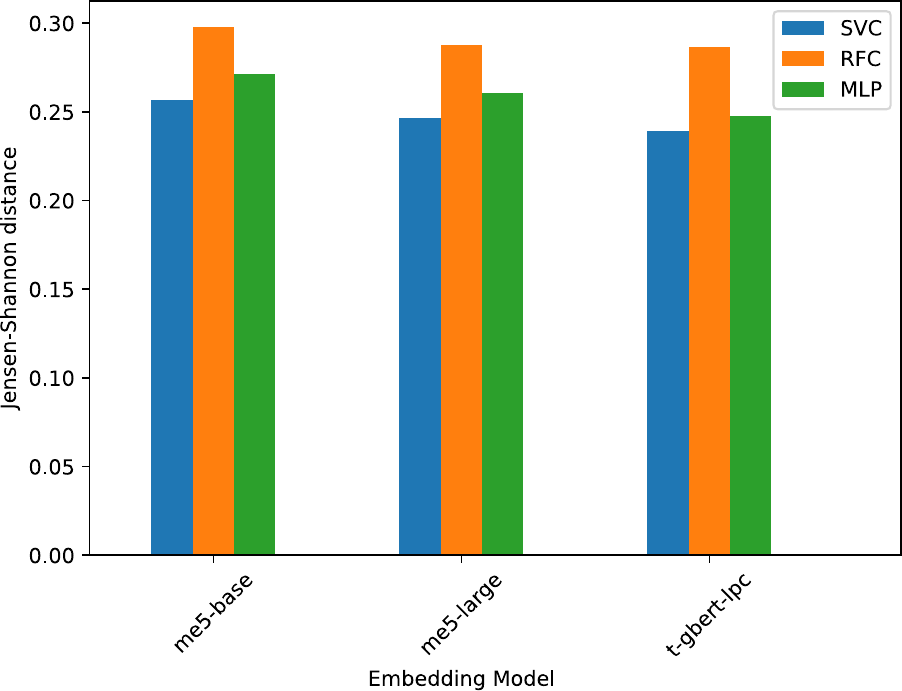}
    \caption{
        Jensen-Shannon distances our models achieved when aggregating the predictions for Subtask 2 on the validation set (lower is better).
    }
    \label{fig:validation_subtask2}
\end{figure}

Figures~\ref{fig:validation_subtask1} and~\ref{fig:validation_subtask2} show the performance of our models on the validation set at the end of the model exploration phase for Subtask 1 and Subtask 2, respectively.
The Multilayer Perceptron and the Support Vector Machine achieved similar scores on both subtasks, with the MLP performing slightly better on Subtask 1 and the SVC on Subtask 2 for all text embedding models.
In comparison, the Random Forest classifier performed worse on average on both subtasks.
While model performance did not vary significantly between the different text embedding models, the mE5-large embeddings consistently outperformed the mE5-base embeddings.
The embeddings from the GBERT-large-pc model performed slightly worse than the mE5-large embeddings on Subtask 1 but achieved slightly better results on Subtask 2.

Accordingly, we decided to submit variations of the Multilayer Perceptron and the Support Vector Machine classifiers with mE5-large and GBERT-large-pc embeddings to the competition.
Our best-performing model for Subtask 1 was the Support Vector Machine classifier on top of mE5-large embeddings with a Macro-F1 score of 0.597, ranking 4th on the Codabench leaderboard.
For Subtask 2, the Support Vector Machine classifier with GBERT-large-pc embeddings achieved the best results with an average Jensen-Shannon distance of 0.301, ranking 2nd on the leaderboard.

\section{Conclusion}

Our study demonstrates that support vector machines trained on open-source text embeddings can robustly predict sexism and misogyny in German-language online comments, reflecting the diversity of human judgments.
While the efficiency and low cost of our training process is an advantage of our approach, we see potential for further improvements.
For example, jointly fine-tuning text embeddings and classifiers on sexism datasets could improve performance, as demonstrated by the top system in the EXIST 2022 challenge, which excelled at detecting sexism in Twitter tweets~\citep{Serrano2022}.
In addition, creating ensembles of fine-tuned text embedding models and classifiers could also lead to better results, similar to the strategy employed by the winning system in the GermEval 2021 challenge for identifying toxic Facebook comments~\citep{Bornheim2021}.
We foresee that advanced sexism detection systems can greatly assist social media moderators, paving the way for more respectful and inclusive online interactions in the future.


\section*{Acknowledgements}
We are grateful to M. Reißel and V. Sander for providing us with computing resources.

\end{document}